\documentclass[letterpaper, 10 pt, conference]{ieeeconf}  % Comment this line out if you need a4paper

\IEEEoverridecommandlockouts                              % This command is only needed if 
\usepackage{mathptmx} % assumes new font selection scheme installed
\usepackage{amsmath} % assumes amsmath package installed
\usepackage{amssymb}  % assumes amsmath package installed
\usepackage{subfig}
\usepackage{mathtools}
\usepackage{float}
\usepackage{hyperref}

\title{\LARGE \bf
BIMRL: Brain Inspired Meta Reinforcement Learning
}

\author{Seyed Roozbeh Razavi Rohani $^{1}$, Saeed Hedayatian $^{2}$, Mahdieh Soleymani Baghshah $^{1}$  
\thanks{$^{1}$Department of Computer Engineering
Sharif University of Technology
        {\tt\small razavii@ce.sharif.edu, soleymani@sharif.edu}}
\thanks{$^{2}$Department of Mathematical Sciences Sharif University of Technology
        {\tt\small hedayatians@gmail.com}}%
}

\begin{document}

\maketitle
\thispagestyle{empty}
\pagestyle{empty}

%%%%%%%%%%%%%%%%%%%%%%%%%%%%%%%%%%%%%%%%%%%%%%%%%%%%%%%%%%%%%%%%%%%%%%%%%%%%%%%%
\begin{abstract}
Sample efficiency has been a key issue in reinforcement learning (RL). An efficient agent must be able to leverage its prior experiences to quickly adapt to similar, but new tasks and situations. Meta-RL is one attempt at formalizing and addressing this issue. Inspired by recent progress in meta-RL, we introduce BIMRL, a novel multi-layer architecture along with a novel brain-inspired memory module that will help agents quickly adapt to new tasks within a few episodes. We also utilize this memory module to design a novel intrinsic reward that will guide the agent's exploration. Our architecture is inspired by findings in cognitive neuroscience and is compatible with the knowledge on connectivity and functionality of different regions in the brain. We empirically validate the effectiveness of our proposed method by competing with or surpassing the performance of some strong baselines on multiple MiniGrid environments.

\end{abstract}

%%%%%%%%%%%%%%%%%%%%%%%%%%%%%%%%%%%%%%%%%%%%%%%%%%%%%%%%%%%%%%%%%%%%%%%%%%%%%%%%
\section{INTRODUCTION}

A major problem with most model-free RL methods is their sample inefficiency and poor generalizability. Even for simple tasks, these methods require thousands of interactions with the environment. Furthermore, even after learning a task, slight changes in the environment dynamics or the underlaying task will force us to basically retrain the agent from scratch. To combat these issues, a number of methods have been proposed that utilize ideas from multi-task and meta learning (\cite{ zintgraf2019varibad, wang2016learning}). Some of these methods treat the task-ID like an unobservable latent variable in the underlaying POMDP and condition their policy on a learned, task belief state \cite{zintgraf2019varibad}. However, these methods generally use some simplifying assumptions that limits their generalizability. Our proposed method, BIMRL, is similar in that it learns to identify a suitable task-specific latent and uses it to quickly adapt, but relaxes some of these assumptions. Similar to some prior works (\cite{zintgraf2019varibad, zintgraf2021exploration}), we use variational methods to estimate this latent variable. However, by introducing a new factorization of the log-likelihood of trajectories, our method learns a different context embedding that is more consistent with POMDP assumptions and can also be interpreted in a more meaningful way.

Another central issue with RL is the well-known exploration-exploitation dilemma \cite{zintgraf2019varibad}. If an agent wants to quickly adapt to a new task, it must be able to explore the environment and obtain relevant experiences. Nonetheless, with complex tasks and environments, conventional exploration methods, such as epsilon-greedy exploration, are not enough. Intrinsic rewards that encourage suitable exploration are currently one of the most successful approaches. Our novel memory module provides an intrinsic reward to guide the exploration; thus alleviating the problems that arise from exploration across different tasks. Aside from this reward our memory module, which consists of an episodic part that resets after each episode, and a Hebbian part (\cite{limbacher2020h}) which retains information across different episodes of a particular task, gives us a performance boost in memory-based tasks while helping deal with catastrophic forgetting that comes up in multi-task settings.

To summarize, our contributions are as follows: 
\begin{enumerate}
    \item
Proposing a multi-layer architecture that bridges model-based and model-free approaches by utilizing predictive information on state-values. Each layer of our architecture corresponds to a different part of the Prefrontal Cortex (PFC) which is known to be responsible for important cognitive functionalities such as learning a world-model \cite{russin2020deep}, planning \cite{miller2021multi}, and shaping exploratory behaviours \cite{russin2020deep}.
One of the objectives in this architecture is derived from our newly proposed factorization of the log-likelihood of trajectories. This factorization is more robust to violations of POMDP assumptions and also takes the predictive coding ability of the brain into consideration \cite{alexander2018frontal}.

    \item
Introducing a neuro-inspired memory module compatible with discoveries in cognitive neuroscience. This module will help in memory-based tasks and give the agent the ability to preserve information, both within an episode and across different episodes of some task. The design of this memory module is reminiscent of Hippocampus (HP) as the main region assigned to episodic memory in the brain \cite{eichenbaum2017prefrontal}.

    \item
Proposing a new intrinsic reward based on our memory module. This reward does not disappear over time, is task-agnostic, and is very effective in a multi-task setting. It will encourage exploratory behaviour that leads to faster task identification.
\end{enumerate}

\section{RELATED WORK}

\subsection{Meta Reinforcement Learning}
Meta reinforcement learning is a promising approach for tackling few-episode learning regimes. It aims to achieve quick adaptation by learning inductive biases in the form of meta-parameters. There are multiple approaches to meta-RL, two of the most popular being gradient based methods and metric based methods.
Another popular family of approaches use a recurrent neural network (RNN) \cite{wang2016learning}. Some previous works try to extend this approach and use an RNN to extract some context that will be used to inform the policy. Recently, by observing the effectiveness of variational approaches for estimating the log-likelihood of trajectories, some methods proposed to use the estimated context (referred to as the belief state in the literature) to inform the policy about its current task (\cite{zintgraf2019varibad, zintgraf2021exploration}). This belief state should contain information about environment dynamics as well as the reward function of the task. Ideally, conditioning on this belief would convert the POMDP into a regular MDP. However, these methods typically impose some constraints on the underlaying POMDP (for instance, \cite{zintgraf2019varibad, zintgraf2021exploration} consider Bayes-adaptive MDPs which are a special case of the general POMDPs). These restrictions may be violated in some scenarios. On the contrary, our proposed method is more robust to such violations.

\subsection{Modular Architectures}
Injecting a modular inductive bias into policies is a promising strategy for improving out-of-distribution generalization in a systematic way.
\cite{goyal2019recurrent} proposed a modular structure where each module can become an expert on a different part of the state space. These modules will be dynamically combined through attention mechanisms to form the overall policy.
\cite{mittal2020learning} introduced an extension called the bidirectional recurrent independent mechanism (BRIMs). BRIM is composed of multiple layers of recurrent neural networks where each module also receives information from the upper layer. A modified version of BRIM is used in our architecture as well.

\subsection{Combining Model-Based \& Model-Free}
There has been efforts on trying to combine model-based and model-free approaches. Dyna \cite{sutton1991dyna} and successor representation \cite{momennejad2017successor} are two such attempts. There has also been some investigations on whether our brains work in a model-based or model-free manner, which has resulted in some neuro-inspired computational models (\cite{kim2020reliability}). We propose a new way of combining these two approaches that is inspired by the predictive capabilities of PFC \cite{russek2017predictive}.

\subsection{Exploration}
Another central issue in RL is efficient exploration of the environment. Methods that encourage effective exploratory behaviour through the use of an intrinsic reward are among the most popular and effective ways of dealing with this problem. Intrinsic rewards can be obtained from a novelty or curiosity criterion (\cite{pathak2017curiosity}). It is worth noting that effective exploration becomes an even bigger issue in the meta-learning setting (\cite{zintgraf2021exploration}).

\section{METHODS}
In this chapter we introduce our proposed method. We will first derive a factorization for a trajectory's log-likelihood. Using this factorization, we will design a multi-layered architecture with multiple loss functions. We will then shift our focus and introduce our memory module. Finally, we will give some intuitive interpretations of our proposed method and mention some of the connections to different regions of the human nervous system. 
\begin{figure*}[!ht]
	\includegraphics[width=\textwidth,height=8cm]{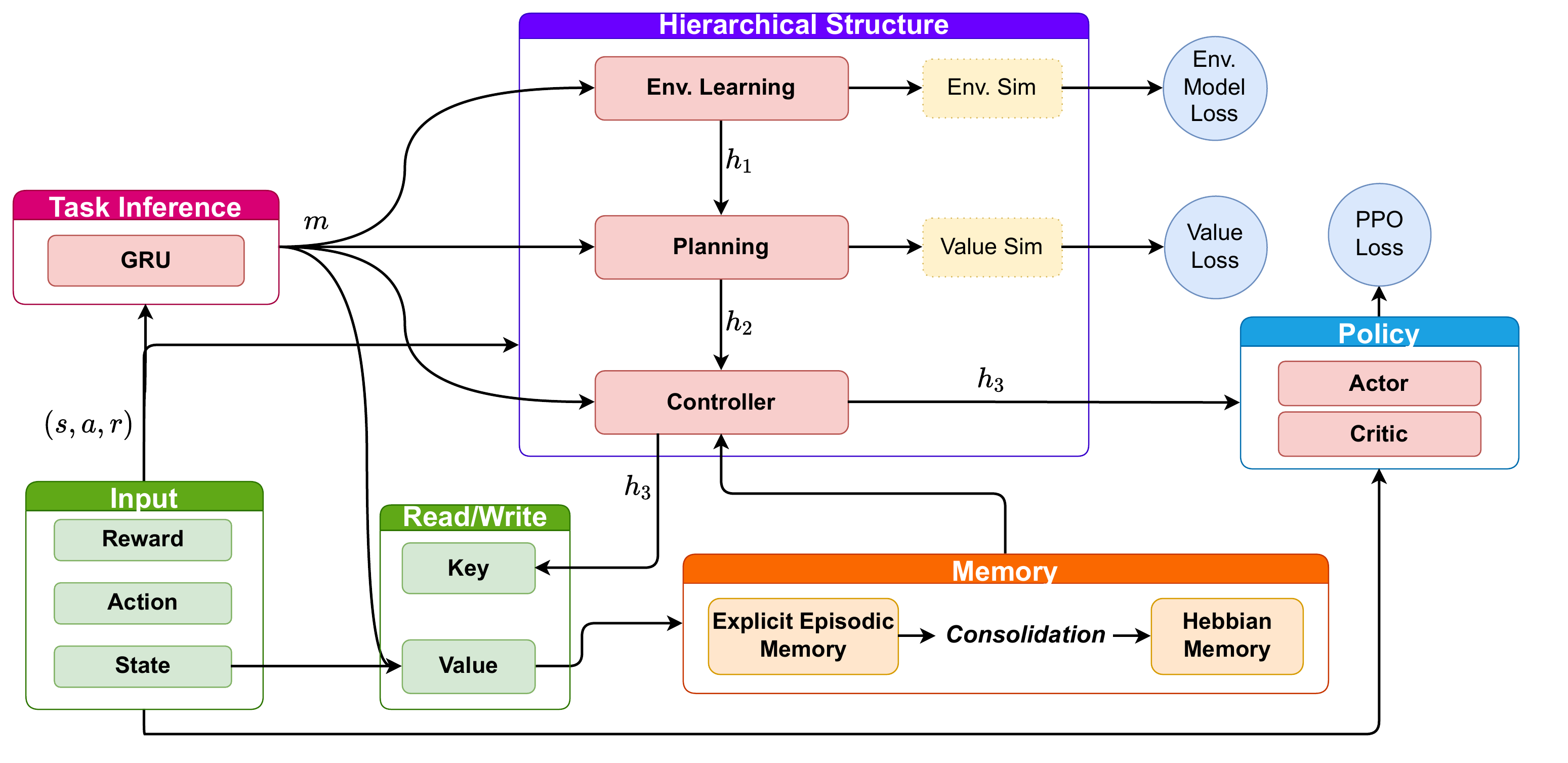}
	\caption{A bird's-eye view of our model. The major components of BIMRL include the task inference module, the multi-layer hierarchical structure and the memory module.}
	\label{fig:full_model}
\end{figure*}
\subsection{Likelihood Factorization}
Our method builds upon VariBAD \cite{zintgraf2019varibad} and we use the same terminology as is used there. Similar to VariBAD, we aim to optimize the following ELBO loss corresponding to log-likelihood of a trajectory $\tau$:
\begin{equation}
	\begin{multlined}[t]
		\mathrm{ELBO}_t = \mathbb{E}_{\rho}\left[\mathbb{E}_{q_{\phi}\left(m \mid \tau_{: t}\right)}\left[\log p_{\theta}\left(\tau_{: H^{+}} \mid m\right)\right] \right. \\ \left.
		-K L\left(q_{\phi}\left(m \mid \tau_{: t}\right) \| p_{\theta}(m)\right)\right],
	\end{multlined}
\end{equation}
where $\rho$ is the trajectory distribution induced by our policy, $m$ is the belief state, and $H^+$ is the time horizon for a task, which is set to be four times the horizon for an episode.
Notice that this objective is comprised of two parts. The first part, $\mathbb{E}_{q_{\phi}\left( m \mid \tau_{: t} \right) } \left[ \log p_{\theta} \left( \tau_{: H^{+}} \mid m \right) \right]$, is known as the reconstruction loss and the second part is the KL-divergence between the variational posterior $q_\phi$ and the prior over the belief state, $p_\theta(m)$. Unlike VariBAD, we factorize the reconstruction loss as follows:
\begin{align} \label{bimrl}
	\log p&\left(\tau_{: H^{+}} \mid m, a_{0,\cdots, n}\right) =\log p\left(s_{0} \mid m\right) \\ &+\sum_{i=0}^{H^{+}-1}\sum_{j=0}^{\text{min}(n, i)}\left[\log p\left(s_{i+1} \mid s_{i-j}, a_{i-j, \cdots, i}, m\right)\right]\nonumber
	\\ &+\sum_{i=0}^{H^{+}-1}\sum_{j=0}^{\text{min}(n, i)}\left[\log p\left(r_{i+1} \mid s_{i-j}, a_{i-j, \cdots, i}, m\right)\right]\nonumber 
	\\ &+\sum_{i=0}^{H^{+}-1}\sum_{j=0}^{\text{min}(n, i)}\left[\log p\left(a_{i+1} \mid s_{i-j, \cdots, i+1}, m\right)\right]\nonumber 
\end{align}
It is worth mentioning that unlike previous methods which use the Bayes Adaptive Markov Decision Process (BAMDP) formulation and assume perfect observability given the belief state (which will not hold in extreme partial observability scenarios), our factorization uses the predictive information in $m$ and asserts that our belief should contain enough information to predict the next $n$ states and rewards, given the next $n$ actions.
Contrary to VariBAD that conditions the trajectory on the full history of actions, we condition a trajectory on only the first $n$ actions (hence, the added fourth summation term in \ref{bimrl}). This modification will result in an added action simulation network which will also help in learning better representations in the first level of the hierarchical structure.
This predictive coding capability is compatible with those observed in the PFC (\cite{alexander2018frontal}).

\iffalse
We can unroll the aforementioned factorization for one time-step and arrive at the following recursive equation. In the next part, we will model these recursive equations with recurrent networks, that we shall call simulation networks, and use them in our proposed architecture.
\begin{align}\label{bimrl_1}
	\log& p\left(\tau_{: H^{+}} \mid m, a_{0\cdots n}\right) =\log p\left(s_{0} \mid m\right) \\ \nonumber &+\sum_{i=0}^{H^{+}-1}\sum_{j=0}^{\text{min}(n, i)}\left[\log p\left(s_{i+1} \mid s_{i}, a_{i}, m\right) + \log p\left(s_{i} \mid s_{i-j}, a_{i-j \cdots i-1}, m\right)\right]
	\\ \nonumber &+\sum_{i=0}^{H^{+}-1}\sum_{j=0}^{\text{min}(n, i)}\left[\log p\left(r_{i+1} \mid s_{i}, a_{i}, m\right) + \log p\left(s_{i} \mid s_{i-j}, a_{i-j \cdots i-1}, m\right)\right]
	\\ \nonumber &+\sum_{i=0}^{H^{+}-1}\sum_{j=0}^{\text{min}(n, i)}\left[\log p\left(a_{i+1} \mid s_{i-j \cdots i+1}, m\right)\right] \nonumber
\end{align}
\fi
\subsection{Brain Inspired Meta Reinforcement Learning (BIMRL)}
Our proposed architecture is composed of a recurrent task inference module responsible for updating our estimate of belief state about the current task, followed by a hierarchical structure similar to BRIM \cite{mittal2020learning}, that conditions on the inferred belief state. As we will see, this hierarchical structure is similar to PFC in how they function. The last layer of this hierarchical structure is a controller whose hidden state will be fed into an actor-critic network. We also have a memory module that directly interacts with the controller. Figure \ref{fig:full_model} demonstrates how all these modules are put together. In what follows, we will briefly introduce each part\footnote{More details about the architectures and hyperparameters that were used as well as the code can be found \href{https://github.com/RoozbehRazavi/BIMRL}{here}}.

%figure 1

\subsubsection{Hierarchical Structure (BRIM)}
The task inference module is a recurrent network that processes the last (observation, action, reward) and produces a belief state that will be passed to each of the three layers in the hierarchical structure as well as the memory module.

\begin{figure}[H]
	\centering
	\includegraphics[width=0.9\linewidth]{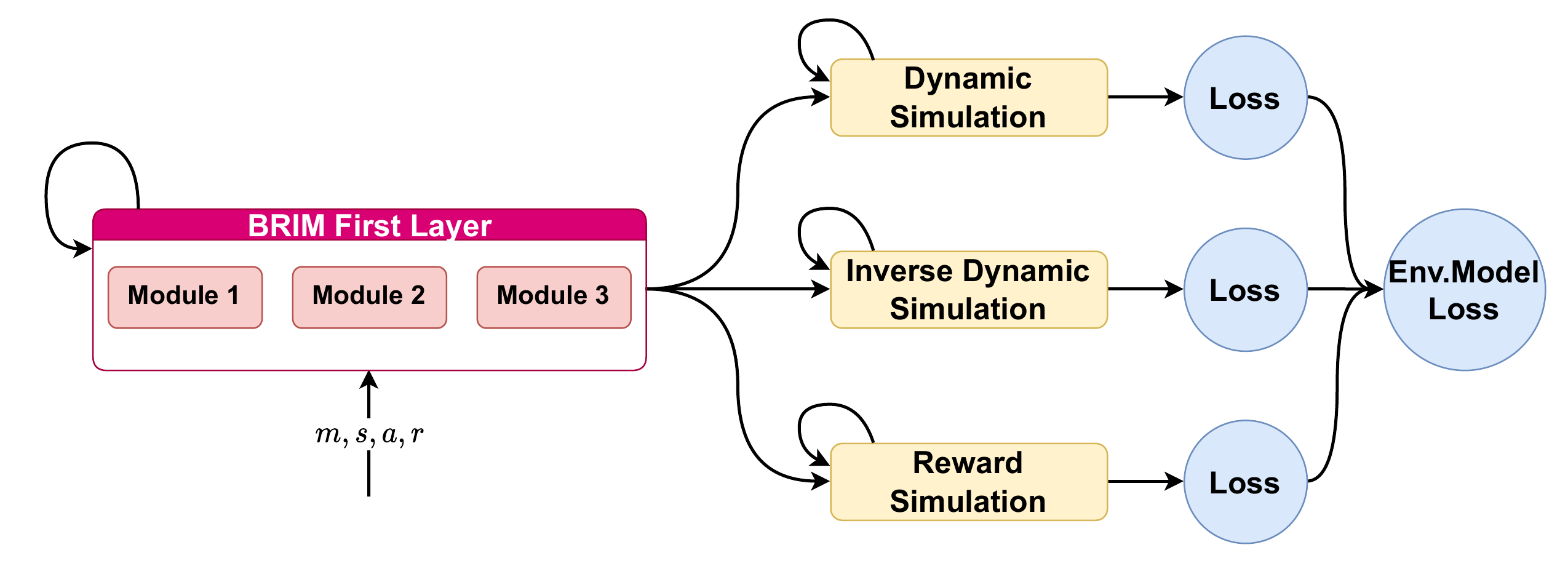}
,	\caption{First level of the hierarchical structure. Responsible for learning a world-model.}
	\label{fig:l1}
\end{figure}

The first level of BRIM (as shown in Fig. \ref{fig:full_model}) is responsible for learning a world model. This layer uses the last observation, action and reward in addition to the current belief state to predict dynamics, inverse dynamics and the rewards. It does so using three simulation networks. This layer is trained to maximize the trajectories log-likelihood as is formulated in \ref{bimrl}. Figure \ref{fig:l1} illustrates this layer in more detail.
The second level (the ``Planning'' box in Fig. \ref{fig:full_model}) is responsible for predicting the value of the next $n$ steps. It will take in the previous level's hidden state ($h_1$) as well as the current state and the next $n$ actions, and predicts the value for the next $n$ time-steps. It aims to minimize the following loss function:
\begin{align}
	Loss_{\mathrm{layer } 2} = \sum_{i=0}^{H^{+}-1}\sum_{j=0}^{n}\left[\left\Vert V_{\psi }\right(s_i, a_{i, \cdots i+j}, h_2\left) - G_{i+j+1: i+j+1+k} \right\Vert^{2}\right], \label{loss_l2} 
\end{align}
where $G_{t:t+k}$ is the $k$-step TD return defined as
\begin{align}
	G_{t: t+k} = R_{t+1}+\gamma R_{t+2}+\cdots+\gamma^{k-1} R_{t+k}+\gamma^{k} V_{\theta}^{\pi}\left(S_{t+k}\right). \label{g_n}
\end{align}
In the above equation, $\psi$ denotes the parameters of $n$-step value decoder network, $h_2$ is the output of the second level and $\theta$ shows the parameters of the critic head in the actor-critic network.

The existence of this layer between the upper layer, responsible for learning the model of the environment, and the lower layer, which directly affects decision making, allows the following interpretations:
\begin{itemize}
	\item \textbf{Model-based approach:}
	Model-based methods first obtain a model of the environment and then plan their next action using this model. In our proposed architecture, the first layer aims to learn as much as possible about the environment. The second layer receives this information and predicts the value for the next $n$ steps. To do so accurately, this layer must be able to perform some sort of planning. Therefore, in our model the two explicit stages in model-based approaches are performed implicitly by the networks in the first and second layer.
	
	\item \textbf{Context-based RL:}
	Methods such as VariBAD inform the agent of its task by providing it with a context vector obtained from a dynamics prediction network. The context vector that we provide is even more informative as it also contains the necessary information for predicting the value of states in the next several steps.
\end{itemize}

Finally, the third level of BRIM, known as the controller, aggregates the information form the previous BRIM layer and the most recent observations form the environment. Additionally, the hidden state of this layer modulates access to the memory, which will be discussed in the next section. The aggregated information will then be used by shallow actor and critic networks to determine the next action and value. We use PPO \cite{schulman2017proximal}, an on-policy learning algorithm, to train our policy networks.
\iffalse
Figure \ref{fig:l3} shows how this layer is connected to the memory module and the actor-critic network.
\begin{figure}[H]
	\centering
	\includegraphics[width=0.75\linewidth]{l3_brim}
	\caption{Third level of the hierarchical structure, i.e., the controller}
	\label{fig:l3}
\end{figure}
\fi

This hierarchical structure is reminiscent of certain parts of the brain.
The first layer is similar to the medial Prefrontal Cortex (mPFC) as they are both responsible for learning a predictive model of the environment \cite{alexander2018frontal}.
Similarly, the second layer corresponds to the Obitofrontal Cortex (OFC) which, in some neuroscientific literature, is identified as the main region responsible for multi-step planning in the brain \cite{miller2021multi}.
Lastly, a part of the nervous system that connects PFC to HP (episodic memory) and the Striatum (which corresponds to actor-critic networks \cite{sutton2018reinforcement}) is known as the Anterior Cingulate Cortex (ACC) \cite{monosov2020outcome}. This region manages the use of control or habitual behaviours based on the degree of uncertainty about the task. It is also known as one of the pathways for transferring information from PFC to HP to guide the transition between the working memory (corresponding to PFC \cite{wang2018prefrontal}) and the episodic memory \cite{eichenbaum2017prefrontal}. In our model this corresponds to the third layer which modulates the memory and aggregates information in time \cite{monosov2020outcome}.

\subsubsection{Memory}
\begin{figure}[!htb]
	\centering
	\includegraphics[width=0.9\linewidth]{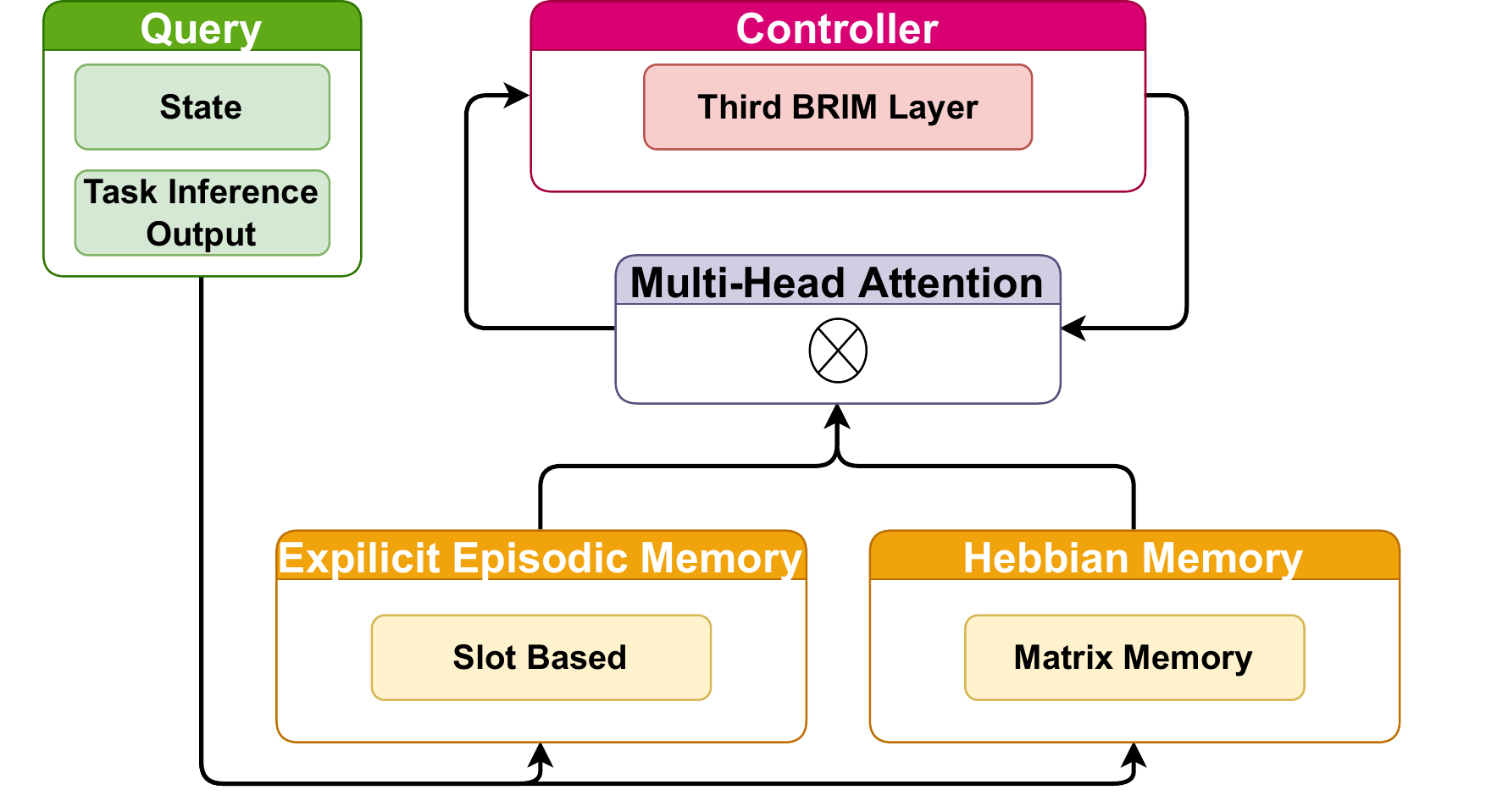}
	\caption{The interaction between the controller and the memory module}
	\label{fig:memory_full}
\end{figure}

The memory module is itself composed of two parts: an explicit episodic memory and a Hebbian memory. At each time step, the output of these two parts are combined using an attention mechanism and is then sent to the controller (see Fig. \ref{fig:memory_full}). The hidden state of the controller is used as the query for this attention mechanism and weights the contribution of each module.

By concatenating the embedded observation of the agent along with the output of the task inference module at each time-step, we create the ``event key'' and the hidden state of the controller is used as the ``event value''. In what follows, we will reference these event keys and values.

\begin{figure}[!htb]
	\centering
	\includegraphics[width=0.9\linewidth]{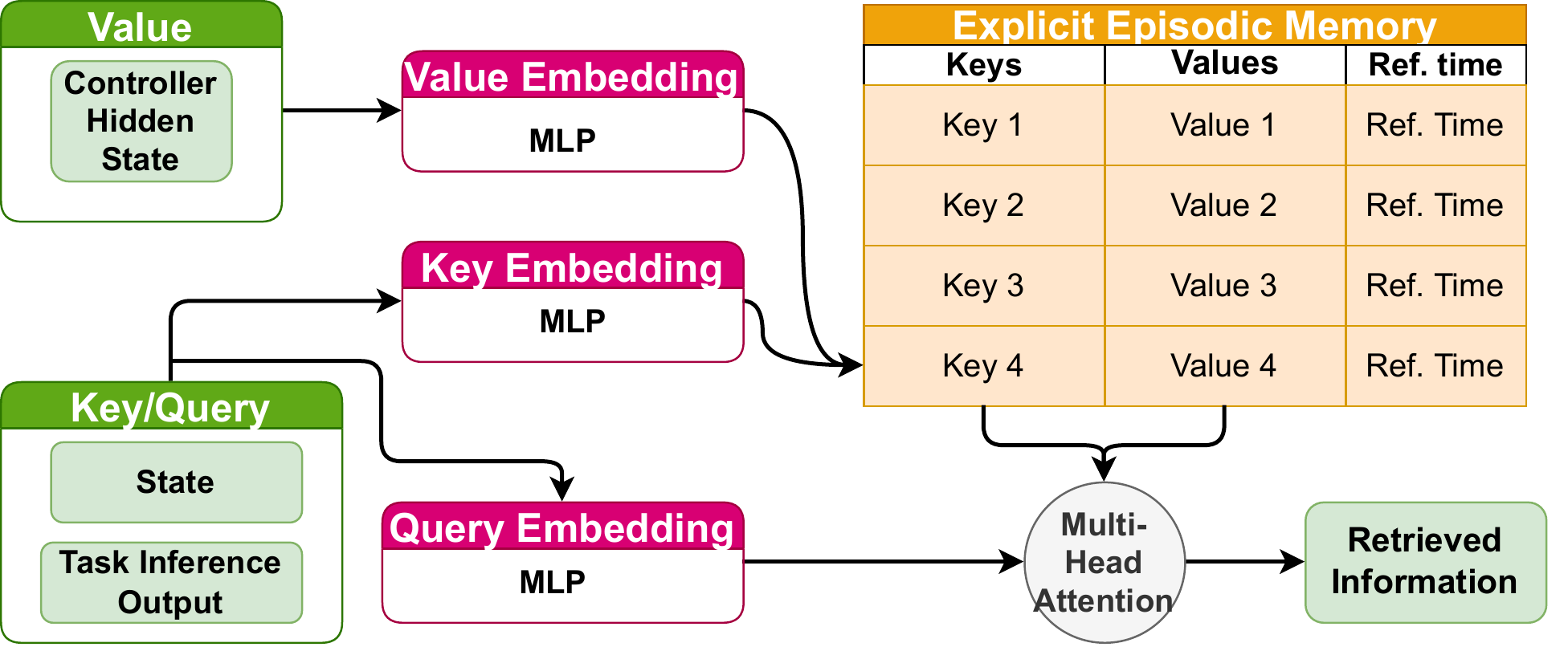}
	\caption{Episodic memory}
	\label{fig:episodic_memory}
\end{figure}

Explicit memory is a slot-based memory that stores the event keys and values. At each time-step, the hidden state of the last level of BRIM is used as the query and the explicit memory is accessed using a multi-head attention mechanism.
At the end of each episode, we compute the normalized ``reference time'' for each slot, which indicates how often each stored event was called upon by the incoming queries, on average. Equation (\ref{equ:ref_time}) shows how it is calculated.
\begin{equation}
    r_{i}=\frac{\sum_{t=s_{i}}^{H} W_{i}^{t}}{H-s_{i}} \label{equ:ref_time}
\end{equation}
In above equation $r_{i}$ is the reference time for the $i$-th slot, $s_{i}$ is the time when this slot was added to the memory, $W_{i}^{t}$ is attention weight attributed to the $i$-th slot at time $t$ and $H$ is the episode length.
As we will see, the reference time is needed for updating the Hebbian memory. Figure \ref{fig:episodic_memory} demonstrates how the episodic memory works.

The Hebbian memory works on the time scale of tasks, i.e., it won't be reset after each episode. Once an episode is completed, the top $k$ percent of slots with the largest reference times are transferred to the Hebbian memory through the process of memory consolidation \cite{camina2017neuroanatomical}. These transferred slots will update the Hebbain memory using the Hebbian learning rule. Equation (\ref{equ:hebb_learning_rule}) shows a general form of this learning rule:
\begin{align}
	\Delta W_{k l}^{\mathrm{assoc}}=\gamma_{+}\left(w^{\mathrm{max}}-W_{k l}^{\mathrm{assoc}}\right) v_{t, k}^{\mathrm{s}} k_{t, l}^{\mathrm{s}}-\gamma_{-} W_{k l}^{\mathrm{assoc}} k_{t, l}^{\mathrm{s}}{ }^{2} \label{equ:hebb_learning_rule}
\end{align}
Here, $\gamma_{+}$ is the correlation coefficient, $\gamma_{-}$ is the regularization coefficient, $w^{\mathrm{max}}$ imposes a soft constraint on the maximum connection weight, and $W_{k l}^{\mathrm{assoc}}$ denotes the parameters of the Hebbian network.
As can be seen in the above equation, this learning rule consists of multiple meta-parameters that control the plasticity of synaptic connections. Inspired by the phenomenon of meta-plasticity (\cite{farashahi2017metaplasticity}), a meta-learning mechanism is used so that these meta-parameters can be learned. Using the Hebbian learning rule as learning mechanism in the inner loop will also circumvent the issue of facing a second order optimization problem.

This transfer of information from a dynamic memory that grows into a fixed-size memory can also help with the decontextualization of the event \cite{duff2020semantic}.

\subsubsection{Intrinsic Reward}
Our proposed intrinsic reward is added to the sparse extrinsic reward of the environment to encourage exploration, both on episodic and task-level time scales. This reward is obtained by multiplying two components. The first component, $r^{\mathrm{curiosity}}_t$, is a curiosity based reward that encourages visiting surprising states. It is a convex combination of reward, state, and action prediction errors. This reward is scaled using a second component, $\alpha_t$. We can think of this second part as a factor that adjusts for the \textit{newness} of the current observation. It is defined as the distance between the most recent observation and the $k$-th nearest event stored in the episodic memory. If the agent encounters an observation that is unlike what has been seen in the on-going episode, this coefficient will be large, resulting in a high intrinsic reward. The following equations fully define the intrinsic reward:
\begin{equation}
    r_{t}^{\mathrm{intrinsic}}\left(r_{t-1}, s_{t-1}, a_{t-1}, m\right)=\alpha_{t}r^{\mathrm{curiosity}}_{t},
\end{equation}
where
\begin{equation}
    \alpha_{t} =  \left\|\boldsymbol{x}_{t}-\mathrm{NN}_{k, \boldsymbol{X}}\left(\boldsymbol{x}_{t}\right)\right\|, \notag
\end{equation}

\begin{align}
    \begin{array}{r}
        r^{\mathrm{curiosity}}_t=-\mathbb{E}_{q_{\phi}\left(m \mid \tau_{: t}\right)}\left[\lambda_{\mathrm{reward}} \times \log p_{\theta}\left(r_{t} \mid s_{t-1}, a_{t-1}, s_{t}, m\right)\right. \\ \notag
        \left.+ \lambda_{\mathrm{state}} \times \log p_{\theta}\left(s_{t} \mid s_{t-1}, a_{t-1}, m\right)\right. \\
        \left.+ \lambda_{\mathrm{action}} \times \log p_\theta (a_t \mid s_t, s_{t-1}, m)\right],
    \end{array}
\end{align}

\begin{align}
	\lambda_{\mathrm{state}} + \lambda_{\mathrm{action}} + \lambda_{\mathrm{reward}} = 1. \notag
\end{align}
Note that $\mathrm{NN}_{k, \boldsymbol{X}}$ is the $k$-th nearest event to the current one, among the records that are stored in the episodic memory, $\boldsymbol{X}$.

\section{EXPERIMENTS}
\begin{figure*}[t]
	\includegraphics[width=\textwidth]{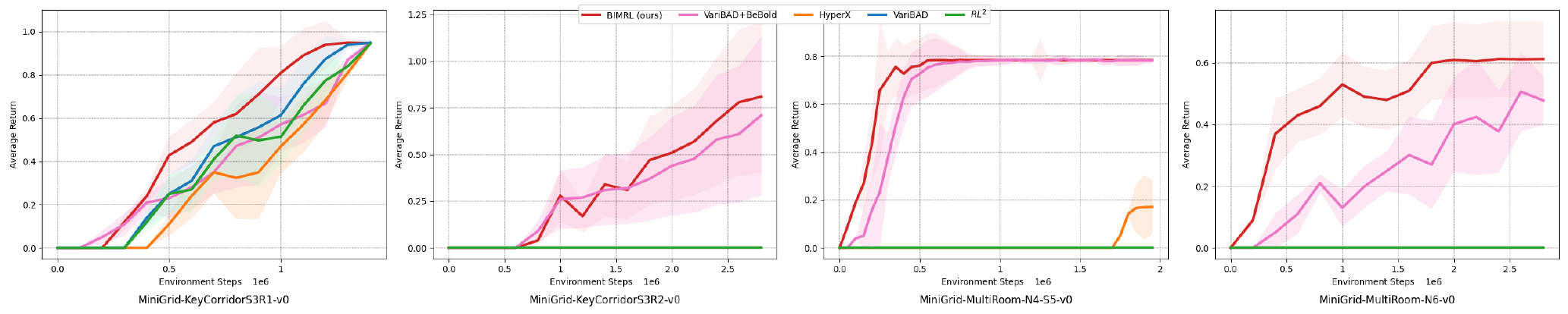}
	\caption{The average return of our proposed model and the baselines in four sets of tasks}
	\label{fig:result}
\end{figure*}
\subsection{Experimental Setup}
For evaluation, we used MiniGrid \cite{gym_minigrid}, a set of procedurally-generated, partially-observable environments in which an agent can interact with multiple objects. At each time-step, the agent receives its observation in the form of a $7\times 7 \times 3$ tensor and can perform any of the available $7$ actions (moving in different directions, picking up objects, etc.). In the meta-training, the agent will see four episodes of each task and has to learn to adapt and solve the task within the time-span of those four episodes. At test time, we report the performance on the last (fourth) episode of each task.
We used four sets of tasks: MultiRoom-N4-S5, MultiRoom-N6, KeyCorridorS3R1, and KeyCorridorS3R2. In the KeyCorridor set of tasks, the agent has to pick-up an object that is behind a locked door. The key is hidden in another room and the agent has to find it. The MultiRoom set of tasks provide an environment with a series of connected rooms with doors between them. The goal is to reach a green square in the final room. 

To compare our results, we used a number of baselines: (\texttt{VariBAD}) \cite{zintgraf2019varibad}, (\texttt{HyperX}) \cite{zintgraf2021exploration}, and (\texttt{Rl\textsuperscript{2}}) \cite{wang2016learning}. We also proposed a new strong baseline by augmenting VariBAD with BeBold \cite{zhang2020bebold} (\texttt{VariBAD+BeBold}), a recently introduced intrinsic reward that improves the performance in sparse-reward environments.
\subsection{Results}
Fig. \ref{fig:result} plots the average return of our model as well as those of our baselines. As can be seen, some of the baselines do not get any reward in the more challenging tasks. This is because those tasks have larger state-spaces and require performing several actions in a particular order. This makes them difficult to solve for methods without a suitable exploration strategy.

In all four sets of tasks, our method achieves the best performance while converging faster and observing fewer number of frames. In MultiRoom-N6, one of the more difficult tasks, our method outperforms VariBAD+BeBold by a significant margin. In this task, other baselines fail to get any reward even after training for more than $2$M frames.

\subsection{Ablation study}
\begin{figure}[H]
	\centering
	\includegraphics[width=0.5 \linewidth,
	height=0.5\linewidth]{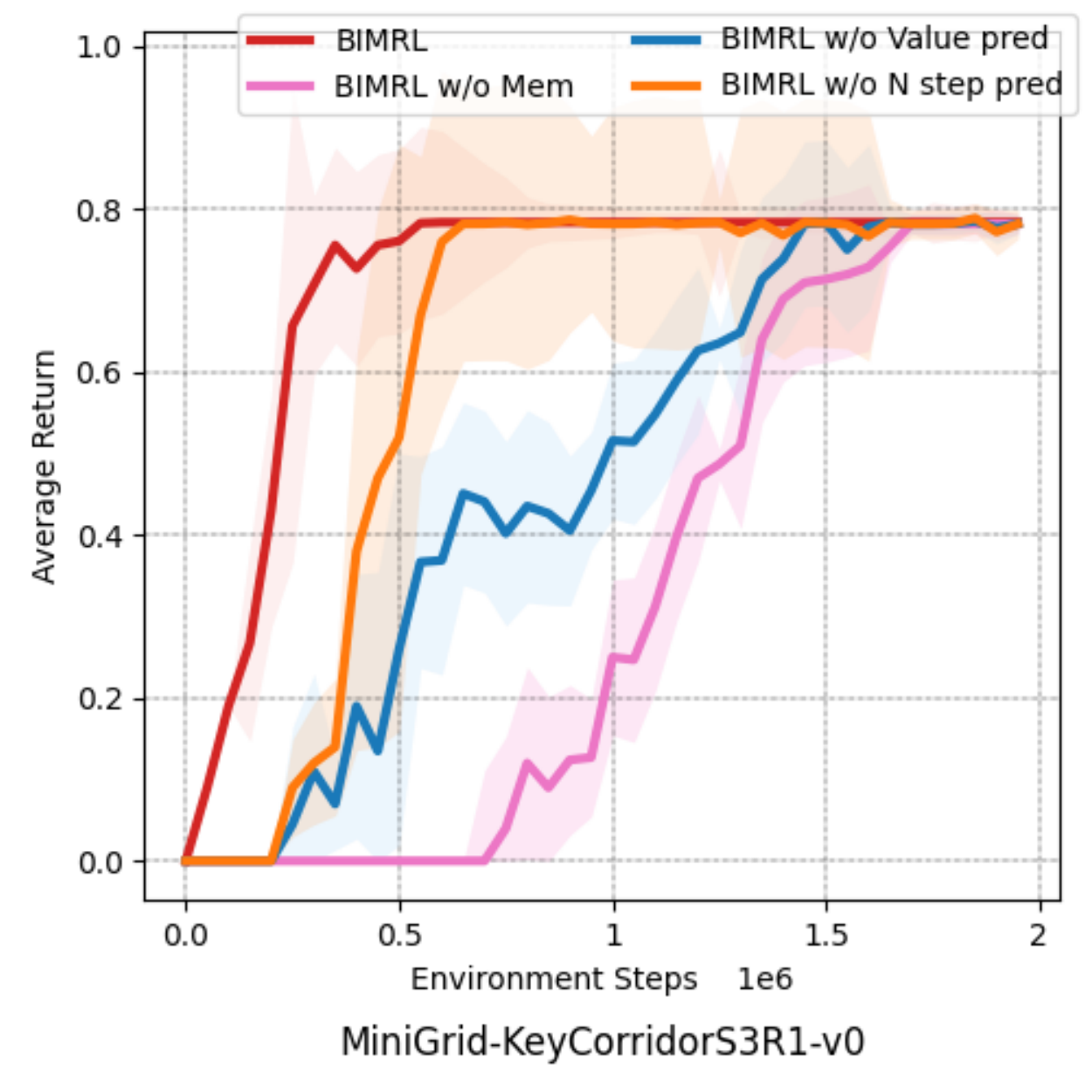}
	\caption{Performance of ablated versions against that of the full model}
	\label{fig:ablation}
\end{figure}

To investigate the significance of each part of our proposed architecture, we evaluated the performance of three ablated versions of our model on the MultiRoom-N4-S5 set of tasks. We examined a model without the memory module (\texttt{BIMRL w/o Mem}), one in which the second level of hierarchical structure (responsible for $n$-step value prediction) was removed (\texttt{BIMRL w/o Value pred}), and one with the vanilla VariBAD trajectory factorization instead of our proposed factorization (\texttt{BIMRL w/o N step pred}).
The result of this ablation study is depicted in Fig. \ref{fig:ablation}. It can be seen that all ablated versions are less sample-efficient, indicating the usefulness of different parts of our model. Most notably, the exclusion of the memory module significantly reduces the convergence rate.

\section{CONCLUSIONS}
We introduced BIMRL, a novel, modular RL agent inspired by the human brain. BIMRL induces several useful inductive biases which helps it meta-learn patterns across different tasks and to adapt quickly to new ones. It emphasizes the significance of taking inspiration from biological systems in designing artificial agents. For future work, one could investigate adding another layer to the memory, in the form of a life-long generative memory module.
Additionally, more extensive experiments on tasks other than those in MiniGrid, particularly memory-based tasks, would shed more light on the extent to which such multi-layered architectures can help with fast adaptation to new situations.
Furthermore, extending our architecture with new modules that make use of textual instructions would allow the agent to test its adaptation capabilities on a much larger set of tasks. This provides yet another avenue for future research.
\addtolength{\textheight}{-5cm}   % This command serves to balance the column lengths

%%%%%%%%%%%%%%%%%%%%%%%%%%%%%%%%%%%%%%%%%%%%%%%%%%%%%%%%%%%%%%%%%%%%%%%%%%%%%%%%

%%%%%%%%%%%%%%%%%%%%%%%%%%%%%%%%%%%%%%%%%%%%%%%%%%%%%%%%%%%%%%%%%%%%%%%%%%%%%%%%

%%%%%%%%%%%%%%%%%%%%%%%%%%%%%%%%%%%%%%%%%%%%%%%%%%%%%%%%%%%%%%%%%%%%%%%%%%%%%%%%
\iffalse
\section*{APPENDIX}

Appendixes should appear before the acknowledgment.

\section*{ACKNOWLEDGMENT}
\fi

%%%%%%%%%%%%%%%%%%%%%%%%%%%%%%%%%%%%%%%%%%%%%%%%%%%%%%%%%%%%%%%%%%%%%%%%%%%%%%%%

\bibliographystyle{IEEEtran}
\bibliography{root}

\end{document}